%% file: root.tex
\documentclass[letterpaper, 10 pt, conference]{ieeeconf}  

\IEEEoverridecommandlockouts                              

\usepackage{graphics} 
\usepackage{graphicx}
\usepackage{amsmath} 
\usepackage{amssymb}  
\usepackage{makecell}
\usepackage{url}
\usepackage{color}
\usepackage{subcaption}
\usepackage{pgfplots}
\usepgfplotslibrary{groupplots,dateplot}
\usetikzlibrary{patterns,shapes.arrows}
\pgfplotsset{compat=newest}

\graphicspath{{pics/}}

\title{\LARGE \bf
Spatial Acoustic Projection for 3D Imaging Sonar Reconstruction
}

\author{Sascha Arnold$^{1,2}$ and Bilal Wehbe$^{1}$
\thanks{$^{1}$German Research Center for Artificial Intelligence, Bremen, Germany}
\thanks{$^{2}$Kraken Robotics, Bremen, Germany\newline
    {\tt\small sarnold@ieee.org, bilal.wehbe@dfki.de}}
}

\begin{document}

\maketitle
\thispagestyle{empty}
\pagestyle{empty}

\begin{abstract}
In this work we present a novel method for reconstructing 3D surfaces using a multi-beam imaging sonar. We integrate the intensities measured by the sonar from different viewpoints for fixed cell positions in a 3D grid.
For each cell we integrate a feature vector that holds the mean intensity for a discretized range of viewpoints. Based on the feature vectors and independent sparse range measurements that act as ground truth information, we train convolutional neural networks that allow us to predict the signed distance and direction to the nearest surface for each cell. The predicted signed distances can be projected into a truncated signed distance field (TSDF) along the predicted directions. Utilizing the marching cubes algorithm, a polygon mesh can be rendered from the TSDF. Our method allows a dense 3D reconstruction from a limited set of viewpoints and was evaluated on three real-world datasets.
\end{abstract}

\section{INTRODUCTION}
Imaging sonars are a key sensor modality for under water vehicles, in particular in higher ranges when optical cameras are limited due to turbid water. 
The reconstruction of 3D information and mapping of the underwater environment is particularly interesting for autonomous underwater vehicles (AUVs) in order to fulfill inspection, exploration and mapping tasks \cite{albiez2015flatfish}.

\begin{figure}
\centering
\begin{subfigure}{.5\columnwidth}
  \centering
  \def\svgwidth{1.1\columnwidth}
  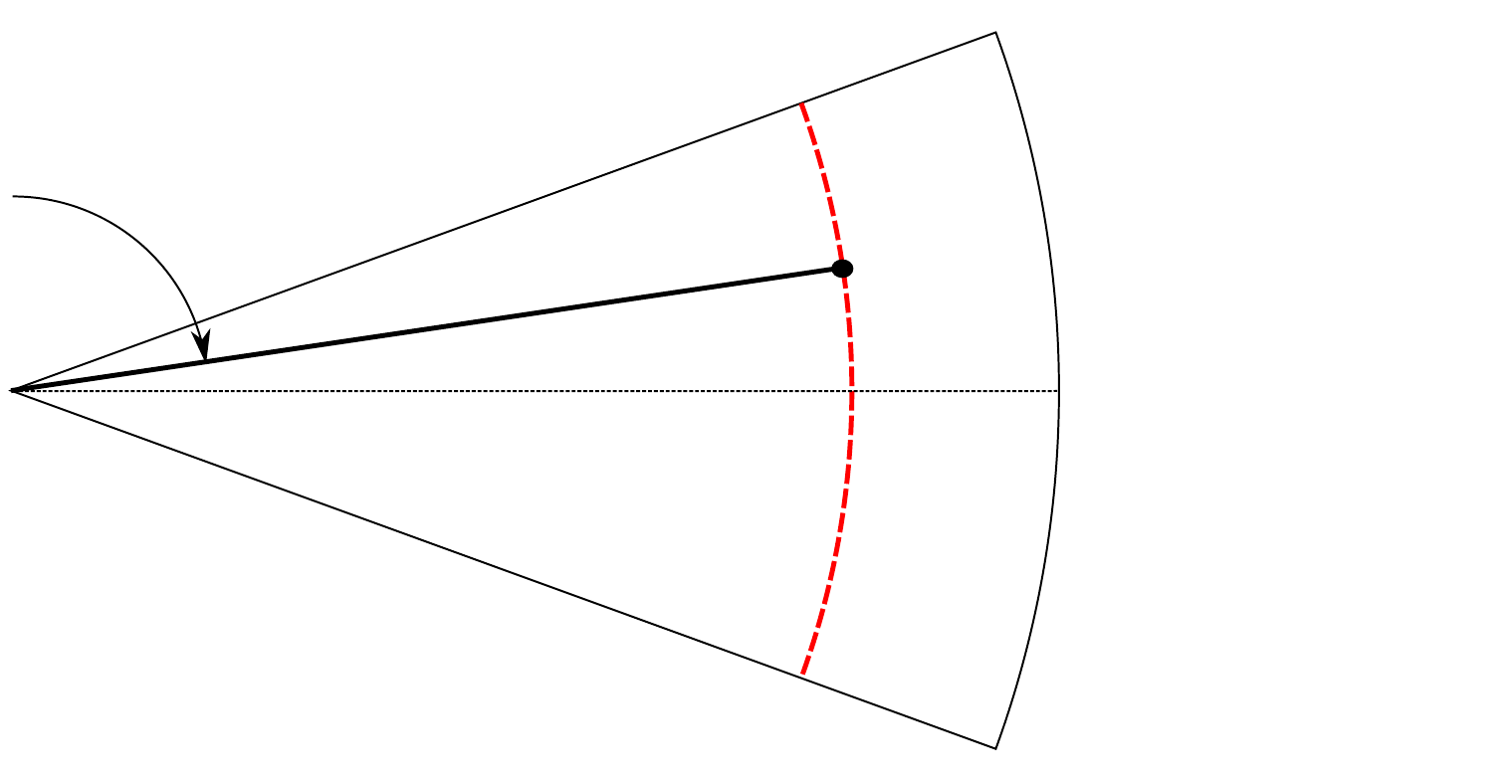
  \caption{The elevation angle $\phi$ (red) is projected into the xy-plane}
  \label{fig:sonar_image_side}
\end{subfigure}%
\begin{subfigure}{.5\columnwidth}
  \centering
  \def\svgwidth{1.1\columnwidth}
  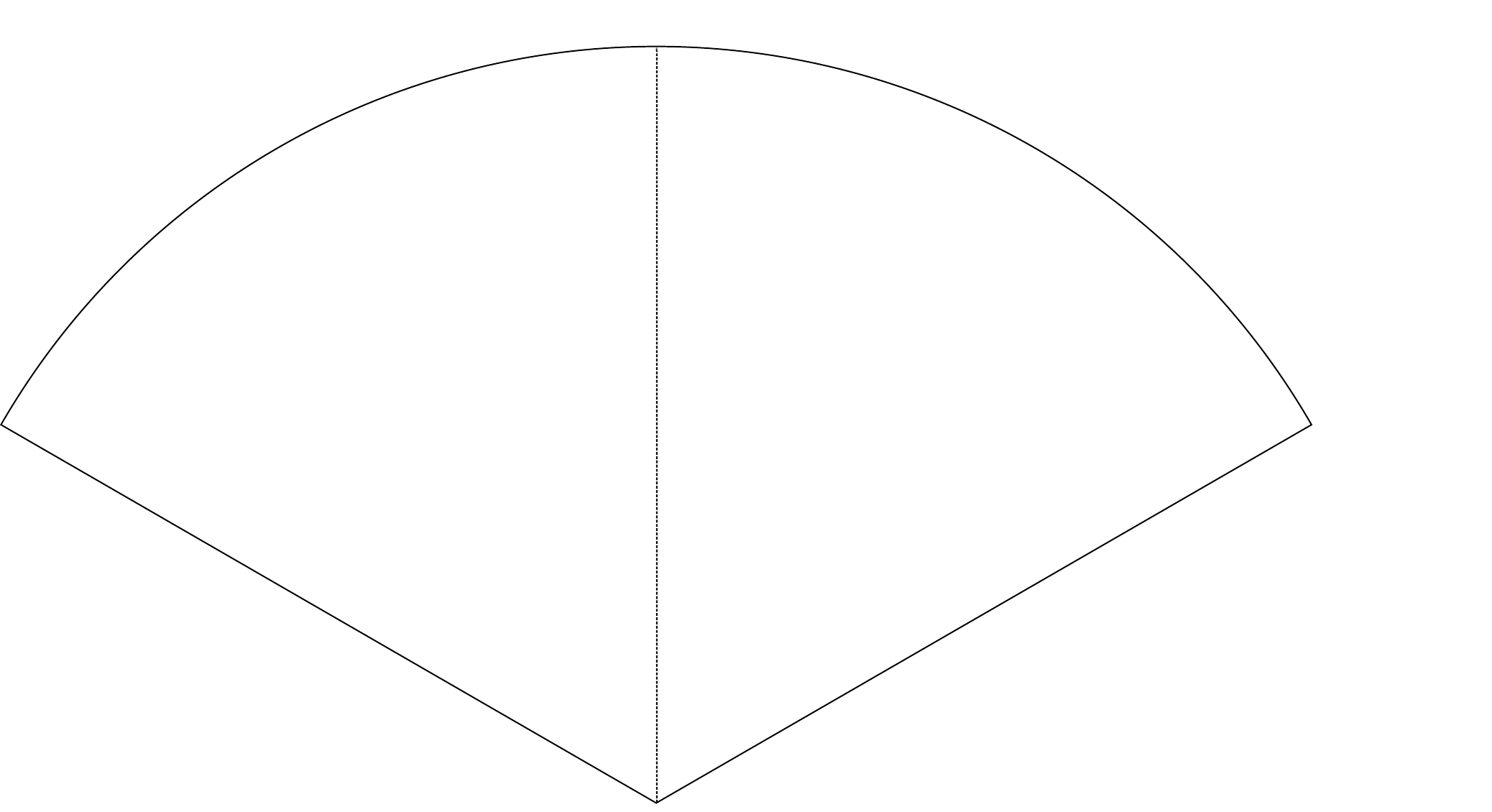
  \caption{Azimuth angle $\theta$}
  \label{fig:sonar_image_above}
\end{subfigure}
\caption{Projection of a 3D point (black dot) into the 2D sonar image $I(r, \theta)$. The elevation angle $\phi$ is lost during acquisition as the 3D point could lie anywhere on the red-dashed arc.}
\label{fig:sonar_image}
\end{figure}

Imaging sonars have a wide vertical opening angle (elevation) allowing to get returns from a subsection of the scene. While the horizontal angle (azimuth) and the range can be measured by the sonar, the elevation angle is lost during acquisition (Fig. \ref{fig:sonar_image_side}). The challenge of 3D imaging sonar reconstruction therefore is to recover the elevation angles to the surfaces in the scene.

Previous works on 3D reconstruction with imaging sonars can be roughly grouped into feature based methods, generative models and volumetric methods.
Feature based methods utilize feature points in the sonar image, and match the corresponding features between images taken from different viewpoints. Based on the feature point correspondences, nonlinear optimization or filtering can be applied to create a sparse 3D reconstruction \cite{huang2015towards, huang2016incremental, wang2019underwater, li2018pose}.
The drawback of those methods is that the reconstruction is sparse, as feature points are often difficult to extract in sonar images and correspondences can be reliably found only at nearby viewpoints.
Generative models utilize the measured intensities and a known starting position in order to derive the slopes of the corresponding surfaces in the scene \cite{aykin2013forward, aykin2016modeling, debortoli2019elevatenet, wang2021elevation, westman2020theory, westman2019wide}. While allowing a locally dense 3D reconstruction, these methods rely on the estimate of object edges and knowledge of the reflective properties of the surfaces.
Volumetric methods discretize the environment into a voxel grid and determine for each voxel their contribution to the sonar image.
Space carving is one example of volumetric methods that utilizes only the free space information before the first high-intensity return in order to carve out the free space, while the remaining voxels are considered occupied \cite{aykin20153, aykin2016three, negahdaripour2017refining, guerneve2015underwater}. Other methods utilize occupancy grid mapping \cite{wang20183d, wang2019three} or more recently albedo-based methods \cite{guerneve2018three, westman2020volumetric}.
These methods rely on a high variety of viewpoints in order to achieve a dense 3D reconstruction.

This work presents a novel volumetric acoustic projection method that can achieve a dense 3D reconstruction with a limited set of viewpoints. As in other volumetric methods we  discretize the environment into a voxel grid, with the difference that every cell stores a feature vector of intensities. Our approach aims to create dense 3D reconstructions with a limited set of viewpoints, for instance if the environment was traversed only once as in a typical survey mission (imaging sonar facing forward and slightly downward).
We model the problem of reconstructing the positions and orientations of surfaces in the scene by defining features with fixed 3D positions that integrate intensities measured from different sensor viewpoints over time. Based on these features, we utilize neural networks in order to first classify outliers and second predict the signed distance and direction to the nearest surface. In order to train the neural networks we rely on the sparse range measurements from a Doppler Velocity Log (DVL) sensor, which act as ground truth information.
The predicted signed distances are projected along their predicted direction into a truncated signed distance field (TSDF). TSDFs have recently become a common implicit surface representation for 3D reconstruction applications \cite{curless1996volumetric, newcombe2011kinectfusion}. Based on the TSDF, a polygon mesh can be rendered using the marching cubes algorithm \cite{lorensen1987marching}. The resulting mesh is used to evaluate our approach against the ground truth information.

Our method relies on the estimated poses provided by an inertial navigation system (INS) in order to relate the different sensor viewpoints. Additionally we utilize the four range measurements provided by a DVL sensor as independent measurements of the surfaces in the scene.



The main contributions of the paper are as follows:
\begin{enumerate}
	\item Novel volumetric acoustic projection method that can predict the signed distance and direction to the nearest surface for each voxel.
	\item Definition of a feature vector for acoustic intensities from different viewpoints.
	\item Show that sparse DVL range measurements can be utilized to train models which allow a dense 3D reconstruction.
\end{enumerate}

In the following section we describe the various aspects of our approach. Section \ref{section:results} presents the results using three real-world datasets. Finally we give conclusion remarks in section \ref{section:conclusion}.

\section{SPATIAL ACOUSTIC PROJECTION}

\begin{figure}
  \centering
  \def\svgwidth{1.0\columnwidth}
  \vspace{5pt}
  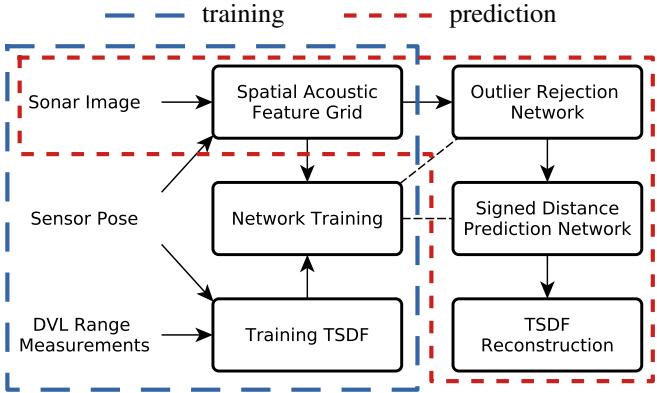
  \caption{Spatial acoustic reconstruction architecture.}
  \label{fig:spatial_acoustic_reconstruction}
\end{figure}

This section details the various aspects of our 3D imaging sonar reconstruction approach. Firstly, a spatial acoustic grid is constructed using the measured intensities by the sonar from different viewpoints. Using the range measurement form a DVL, a TSDF is constructed which acts as a ground truth. Two neural networks are trained in order to relate the feature vectors from the spatial grid with a ground truth signed distance and direction to the nearest surface. The trained networks are then deployed to predict the signed distances and directions creating a TSDF that is used to reconstruct a 3D mesh. A block diagram of the overall architecture is illustrated in Fig. \ref{fig:spatial_acoustic_reconstruction}.

\subsection{Imaging sonar sensor}

An imaging sonar is an active acoustic sensor which emits a pulse of sound and measures the intensity $I$ of the reflected pulse by the scene.
The 2D image of a multi-beam sonar $I(r, \theta)$ is described by the range $r$, which is determined by the speed of sound in water and by the azimuth angle $\theta$, which is determined by a 1D array of transducers. The elevation angle $\phi$ is lost during the projection from the 3D world into the 2D image. 
Therefore, the measured intensity $I(r, \theta)$ includes all reflections along the elevation arc ($\phi_{\textrm{min}}$, $\phi_{\textrm{max}}$) defined by the vertical opening angle of the imaging sonar.
Fig. \ref{fig:sonar_image} illustrates the projection of a point in Euclidean coordinates into the sonar image.


\subsection{Spatial Acoustic Feature Grid}
\label{section:inverse_projection}
We discretize the 3D Euclidean space into a 3D volumetric grid with a fixed cell resolution along each axis.
Each grid cell stores the measured intensity $I$ for each angle between the vector $\mathbf{\hat{r}}_t$ and each axis vector.
$\mathbf{\hat{r}}_t$ points from the grid cell to the sonar for a measurement at time $t$:
\begin{equation}
\mathbf{\hat{r}}_t = \dfrac{(\mathbf{p}_{s,t} - \mathbf{p}_c)}{\lVert(\mathbf{p}_{s,t} - \mathbf{p}_c)\rVert}
\end{equation}
where $\mathbf{p}_{s,t}$ is the position of the sonar at time $t$ and $\mathbf{p}_c$ is the center of the cell $c$.

Given $\mathbf{\hat{r}}_t$, the angles $\gamma_{i}$ to each unit axis $\mathbf{\hat{e}}_{i}$ in the range $[0,\pi/2]$ for $i \in [x,y,z]$ are defined as:
\begin{equation}
\label{gamma}
\gamma_{i,t} = \arccos(\lvert\mathbf{\hat{e}}_{i} \cdot \mathbf{\hat{r}}_t\rvert)
\end{equation}
with 
\begin{equation}
\label{unit}
\lVert\mathbf{\hat{r}}_t\rVert = \lVert\mathbf{\hat{e}}_{i}\rVert = 1 ~\forall i,t
\end{equation}

We limit the angular range to $[0,\pi/2]$ since we assume the same intensity characteristic above and below the surface.

\begin{figure}
  \centering
  \def\svgwidth{1.2\columnwidth}
  \vspace{5pt}
  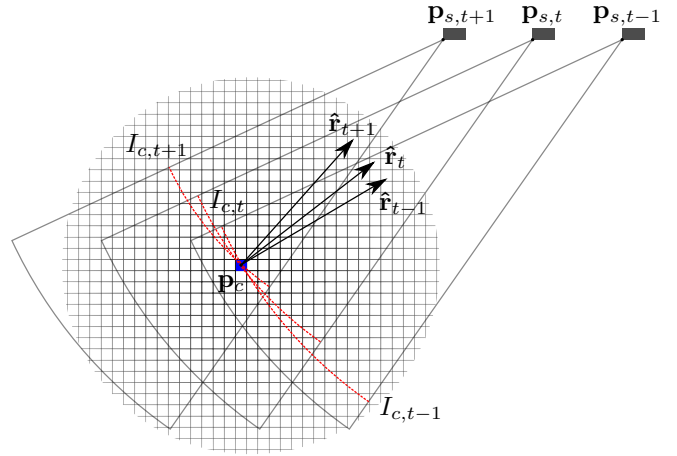
  \caption{Projection of measured intensities into a cell $c$ of the 3D grid from three different sensor viewpoints. The red-dashed arcs visualize the intensity measurement along the elevation angle for each viewpoint.}
  \label{fig:inverse_projection}
\end{figure}

Fig. \ref{fig:inverse_projection} illustrates the projection of three intensity measurements into the same cell from different sensor viewpoints. The intensity $I_{c,t}$ for a measurement at time $t$ at the position of a cell $c$ can be expressed in the sonar image $I_t(r, \theta)$ as:
\begin{equation}
\label{intensity}
I_{c,t} =  I_t(\lVert\mathbf{p}_l\rVert, \arctan(y_l/x_l))
\end{equation}
where $\mathbf{p}_l$ is the cell center $\mathbf{p}_c$ expressed in the sonar frame:
\begin{equation}
\mathbf{p}_l = \begin{bmatrix}
           x_l \\
           y_l \\
           z_l
         \end{bmatrix} = \mathbf{C}^s_{c,t} \mathbf{p}_c
\end{equation}
and $\mathbf{C}^s_{c,t}$ is the coordinate transformation from cell to sonar frame at time $t$.

Based on (\ref{gamma}) and (\ref{intensity}) we define the feature vectors $\mathbf{x}_{i}$ as:
\begin{equation}
\label{feature_vector}
\mathbf{x}_{i}(\gamma_{i,t}) = \dfrac{1}{(T-1)} \sum_{t=0}^{T} I_{c,t}, \, \forall i \in [x,y,z]
\end{equation}
where the range of the angles $\gamma_{i}$ is discretized to a fixed resolution $n$.
A feature vector $\mathbf{x}_{i}$ therefore is of size $n \times 1$.
Each measured intensity $I_{c,t}$ at the cell position $\mathbf{p}_c$ is integrated in each feature vector $\mathbf{x}_{i}$. 
The three feature vectors $\mathbf{x}_{i}$ are arranged into the feature vector $\mathbf{X}$ as follows:
\begin{equation}
\label{X}
\mathbf{X} = [\mathbf{x}_{z}, \mathbf{x}_{y}, \mathbf{x}_{x}]^T
\end{equation}

\begin{figure}
  \centering
  \includegraphics [width=\columnwidth] {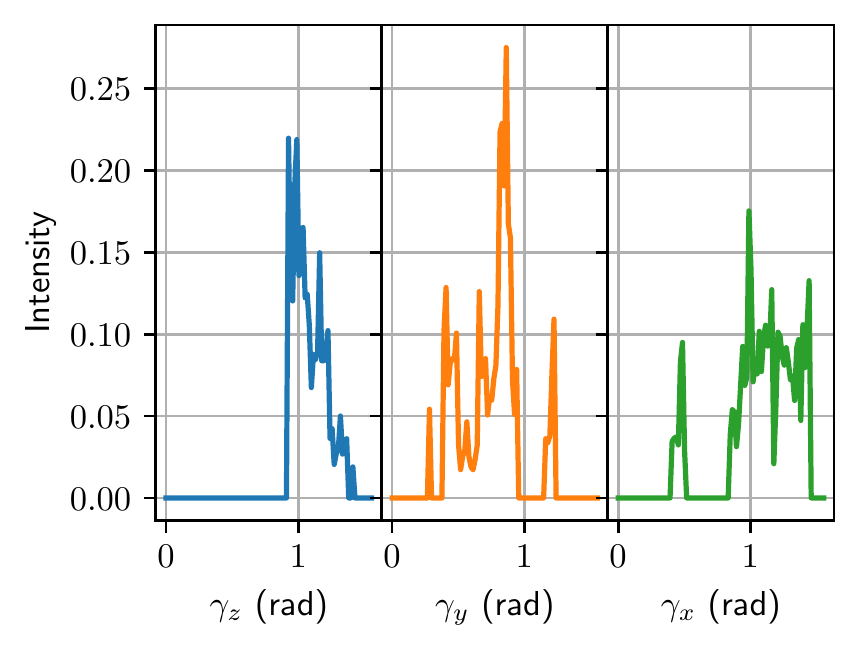}
  \caption{Feature vector $\mathbf{X}$ split into its three components $[\mathbf{x}_{z}(\gamma_{z}), \mathbf{x}_{y}(\gamma_{y}), \mathbf{x}_{x}(\gamma_{x})]^T$.}
  \label{fig:feature_vector_plot}
\end{figure}

$\mathbf{X}$ therefore is of size $3n \times 1$ and will be the input of the neural networks. Each cell $c$ in the 3D grid has a feature vector $\mathbf{X}_c$.
Fig. \ref{fig:feature_vector_plot} shows an example of $\mathbf{X}$ split into its three components $\mathbf{x}_{i}$.

\subsection{Training TSDF}
\label{section:sparse_training_sampels}
We build a TSDF based on the range measurements of a DVL sensor in order to determine a sparse ground truth of the true shape of the environment that can be utilized to train the neural networks.
The range measurements are projected into the TSDF by tracing along the measurement from the direction of the sensor origin around the margin defined by a truncation threshold $\tau$, similar to \cite{newcombe2011kinectfusion}.
By that each visited cell of the TSDF holds the truncated signed distance $d_c$ to the nearest surface. In order to account for the opening angle of each DVL beam we trace along 9 rays, one in the center and 8 equidistantly placed along the outside cone of the DVL beam defined by the opening angle.

To determine the gradients for each cell we apply a 3D Sobel filter with a kernel size of $3 \times 3 \times 3$. This gives us the distance gradients $\mathbf{n}_c$ for each cell with a defined neighborhood.
Since we are interested in values close to zero and a compact representation as model output, we represent the direction as a delta on the unit sphere with respect to the unit vector $\mathbf{\hat{e}}_{z}$.
(\ref{log_map}) computes the logarithm map that maps $\mathbf{n}_c$ to the tangent plane $\log_{\mathbf{\hat{e}}_{z}}(\mathbf{n}_c)$ determined by $\mathbf{\hat{e}}_{z}$ \cite{hertzberg2013integrating}:
\begin{equation}
\label{log_map}
\delta\mathbf{n}_c = \log_{\mathbf{\hat{e}}_{z}}(\dfrac{\mathbf{n}_c}{\lVert\mathbf{n}_c\rVert}) \in \mathbb{R}^2
\end{equation}
The gradient $\mathbf{\hat{n}}_c$ with unit length can be recovered using the exponential map:
\begin{equation}
\label{exp_map}
\mathbf{\hat{n}}_c = \exp_{\mathbf{\hat{e}}_{z}}(\delta\mathbf{n}_c) \in \mathbb{S}^2
\end{equation}
With the tuple $[d_c, \delta\mathbf{n}_c]^T$ the signed distance and direction to the nearest surface can be described for each cell $c$.

\subsection{Neural network architectures}
\label{section:nn_architecture}
We are using two cascaded neural networks, the first one is a classifier and the second is a regressor. The outlier rejection network predicts for each feature vector if the corresponding cell is inside of the truncation threshold to the nearest surface. The signed distance prediction network predicts for positively classified feature vectors the signed distance to the nearest surface and the distance gradient.

The outlier rejection network is a classifier that predicts for each feature vector $\mathbf{X}_c$ if the corresponding cell $c$ is within the truncation threshold $\tau$ to the nearest surface. The boolean training samples $\mathbf{y}_b$ are defined based on the signed distance $d_c$:
\begin{equation}
\label{y_class}
\mathbf{y}_b = [\lvert d_c \rvert < \tau] \in [0,1]
\end{equation}

\begin{figure}
  \centering
  \def\svgwidth{0.96\columnwidth}
  \vspace{5pt}
  \footnotesize
  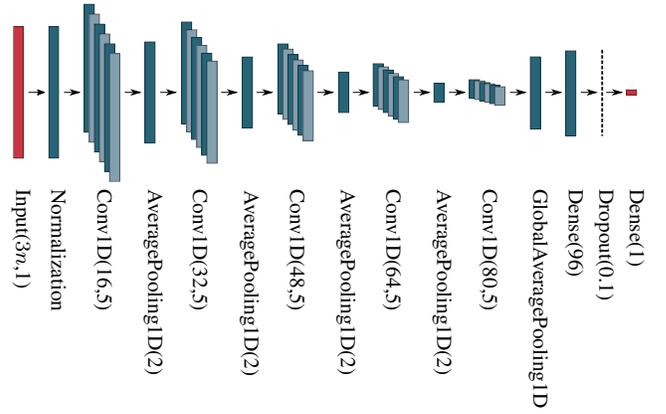
  \caption{Architecture of the outlier rejection network.}
  \label{fig:classifier_network}
\end{figure}

The outlier classification is modeled as a convolutional neural network (CNN) as shown in Fig. \ref{fig:classifier_network} with 5 convolutional layers and 2 fully connected layers at the end. As activation function the rectified linear unit is used in all except the last layer, which uses the sigmoid activation function.
We use a global average pooling layer instead of a flattening in order to reduce the number of trainable parameters. With a feature vector resolution of $n=100$ the network consists of $59393$ trainable parameters.

\begin{figure}
  \centering
  \def\svgwidth{0.96\columnwidth}
  \vspace{5pt}
  \footnotesize
  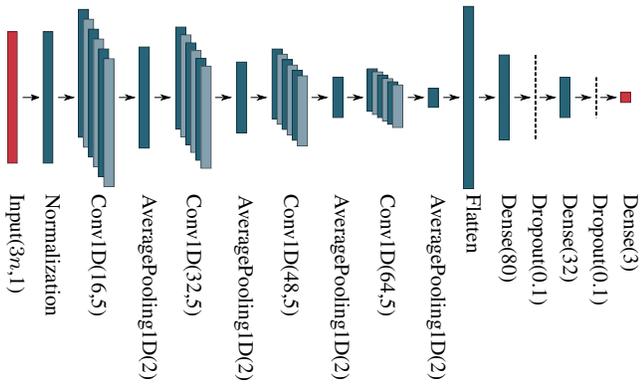
  \caption{Architecture of the signed distance prediction network.}
  \label{fig:regressor_network}
\end{figure}

The signed distance prediction network is a regressor that predicts for each feature vector $\mathbf{X}_c$ of the corresponding cell $c$ the signed distance to the nearest surface and the distance gradient:
\begin{equation}
\label{y_reg}
\mathbf{y}_d = [d_c, \delta\mathbf{n}_c]^T \in \mathbb{R}^3
\end{equation}
The signed distance prediction is modeled as a CNN as shown in Fig. \ref{fig:regressor_network}. It has 4 convolutional layers, a flatten layer and 3 fully connected layers at the end. As activation function the rectified linear unit is used in all convolutional and dense layers. 
With a feature vector resolution of $n=100$ the network consists of $105411$ trainable parameters.

The CNN architectures were selected in an progressive fashion, starting from a simple architecture and adding layers and/or cells as long as the performance improved.



\subsection{TSDF Reconstruction}
\label{section:reconstruction}

As seen in the overview in Fig. \ref{fig:spatial_acoustic_reconstruction} the TSDF reconstruction is the last step of our approach that utilizes the outputs of the neural networks in order to reconstruct a polygon mesh based on the measurements of the imaging sonar.

For each cell in the spatial acoustic feature grid the trained models can predict, using the feature vector $\mathbf{X}_c$ as input, if a cell is inside of the truncation threshold to the nearest surface $\tilde{\mathbf{y}}_b$ and if this is true the signed distance and distance gradient $\tilde{\mathbf{y}}_d$.

Based on the predictions $\tilde{\mathbf{y}}_d$ of the signed distance prediction network, the TSDF is built by tracing along the predicted gradient $\tilde{\mathbf{\hat{n}}}_c$ with the predicted distance $\tilde{d}_c$ in the range of $[\tilde{d}_c-\tau,\tilde{d}_c+\tau]$ using the center of cell $c$ as origin. 
This process is repeated for all cells.
The TSDF and the spatial acoustic feature grid share the same cell resolution.

Using the well known marching cubes algorithm \cite{lorensen1987marching} a polygon mesh can be extracted from the TSDF.


\section{RESULTS}
\label{section:results}

For the evaluation of our method we utilize three datasets that have been collected in the open ocean using the \textit{FlatFish} AUV \cite{albiez2015flatfish}. The AUV is equipped with a Tritech Gemini 720i Multibeam Imaging Sonar and an INS described in \cite{arnold2018robust}. The estimated pose of the AUV during the experiments is affiliated with an growing error throughout the mission \cite{arnold2018robust}. Since in our experiments we are only integrating a limited time window of measurements we are disregarding the pose error and assume the pose to be known.
From the three datasets one was split into a training and evaluation part, while the other two datasets were used only for evaluation.
In all experiments a truncation threshold of $\tau=1.0$ and a feature vector resolution of $n=100$ was selected.

\subsection{Network Training}

\begin{figure}
\centering
\begin{subfigure}{.5\columnwidth}
  \centering
  \includegraphics [width=\textwidth] {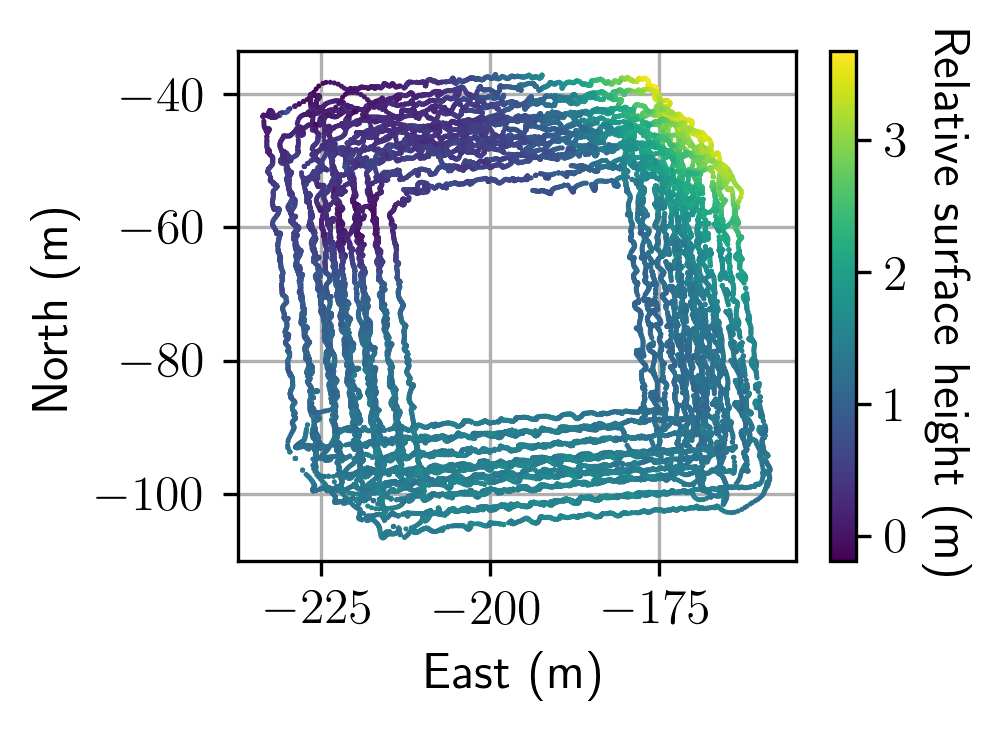}
  \caption{Relative bathymetry height \vspace{4pt}}
  \label{fig:dvl_square_height}
\end{subfigure}%
\begin{subfigure}{.5\columnwidth}
  \vspace{3pt}
  \centering
  \includegraphics [width=\textwidth] {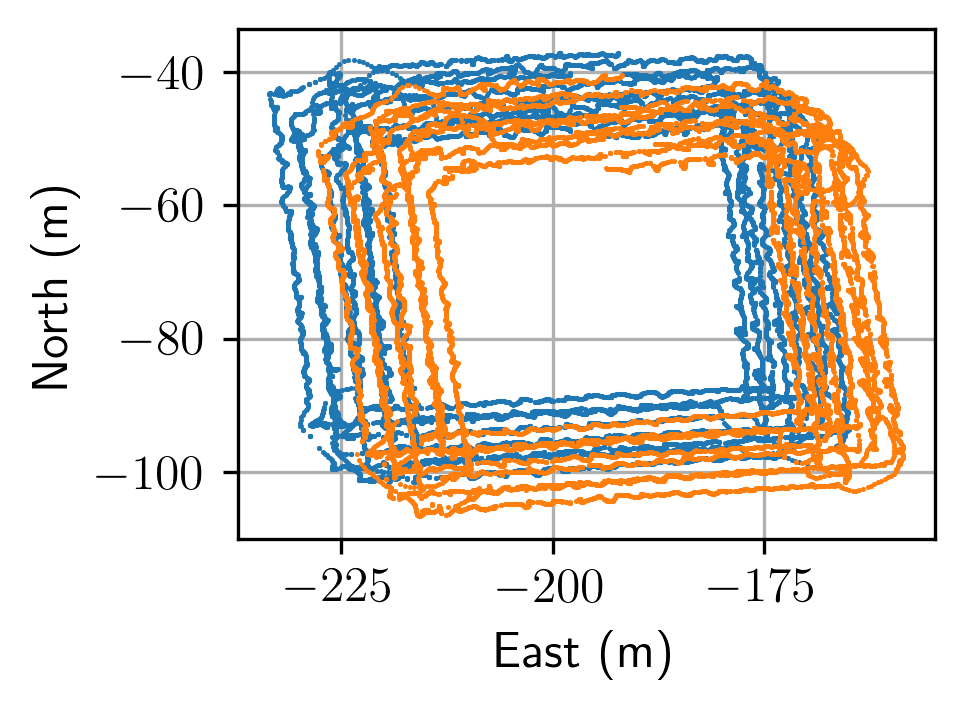}
  \caption{Measurements used for training (blue) and evaluation (orange)}
  \label{fig:dvl_square_split}
\end{subfigure}
\caption{Accumulated DVL range measurements of the square trajectory dataset.}
\label{fig:dvl_square_dataset}
\end{figure}

Fig. \ref{fig:dvl_square_dataset} shows the sparse DVL range measurements of the dataset that was used to train the networks defined in \ref{section:nn_architecture}.
During the mission the vehicle followed repeated square trajectories with an edge length of 50 meter. As can be seen in Fig. \ref{fig:dvl_square_height} the trajectories have a displacement, which allows to cover a wider area of the bathymetry with the four DVL range measurements.
The networks have been trained using the measurements of the first half while the measurements of the second half have been used for the evaluation of the resulting surface reconstruction (Fig. \ref{fig:dvl_square_split}).

The networks are trained on the associated data for each grid cell $c$ in the spatial acoustic feature grid (section \ref{section:inverse_projection}) and the training TSDF (section \ref{section:sparse_training_sampels}). Both 3D grids therefore have the same resolution during the training step.
The outlier rejection network is trained with the feature vector $\mathbf{X}_c$ as input and the boolean class $\mathbf{y}_{b,c}$ as output.
The signed distance prediction network is trained with the same input and the vector $\mathbf{y}_{d,c}$ as output.

The training was performed using the Adam optimizer with a learning rate of $0.0001$, a validation split of $0.2$ and a grid cell resolution of $0.1 m$. As loss the logarithm of the hyperbolic cosine was used for the signed distance prediction network and the binary cross-entropy for the outlier rejection network.
The training was performed until the validation loss was stable for several epochs. The outlier rejection network reached an accuracy of $0.957$ after $85$ epochs and the signed distance prediction network a mean squared error (MSE) of $0.0267$ after $204$ epochs. The selected CNN architectures were outperforming networks utilizing only fully connected layers with a similar amount of trainable parameters (Accuracy: $0.952$, MSE: $0.0415$).

\subsection{Evaluation}

For the evaluation of the 3D reconstruction the second half of the square trajectory dataset, a dataset with a lower altitude to the seafloor and a dataset with man-made structures on the seafloor were used. All reconstructions are preformed based on the models trained with the first half of the square trajectory dataset.

The result of the 3D reconstruction described in section \ref{section:reconstruction} is a polygon mesh of the surfaces in the scene based only on the imaging sonar intensity measurements from the viewpoints defined by the trajectory of the AUV.

\subsubsection{Square trajectory dataset}
\begin{figure}[t!]
  \centering
  \vspace{5pt}
  \includegraphics [width=\columnwidth] {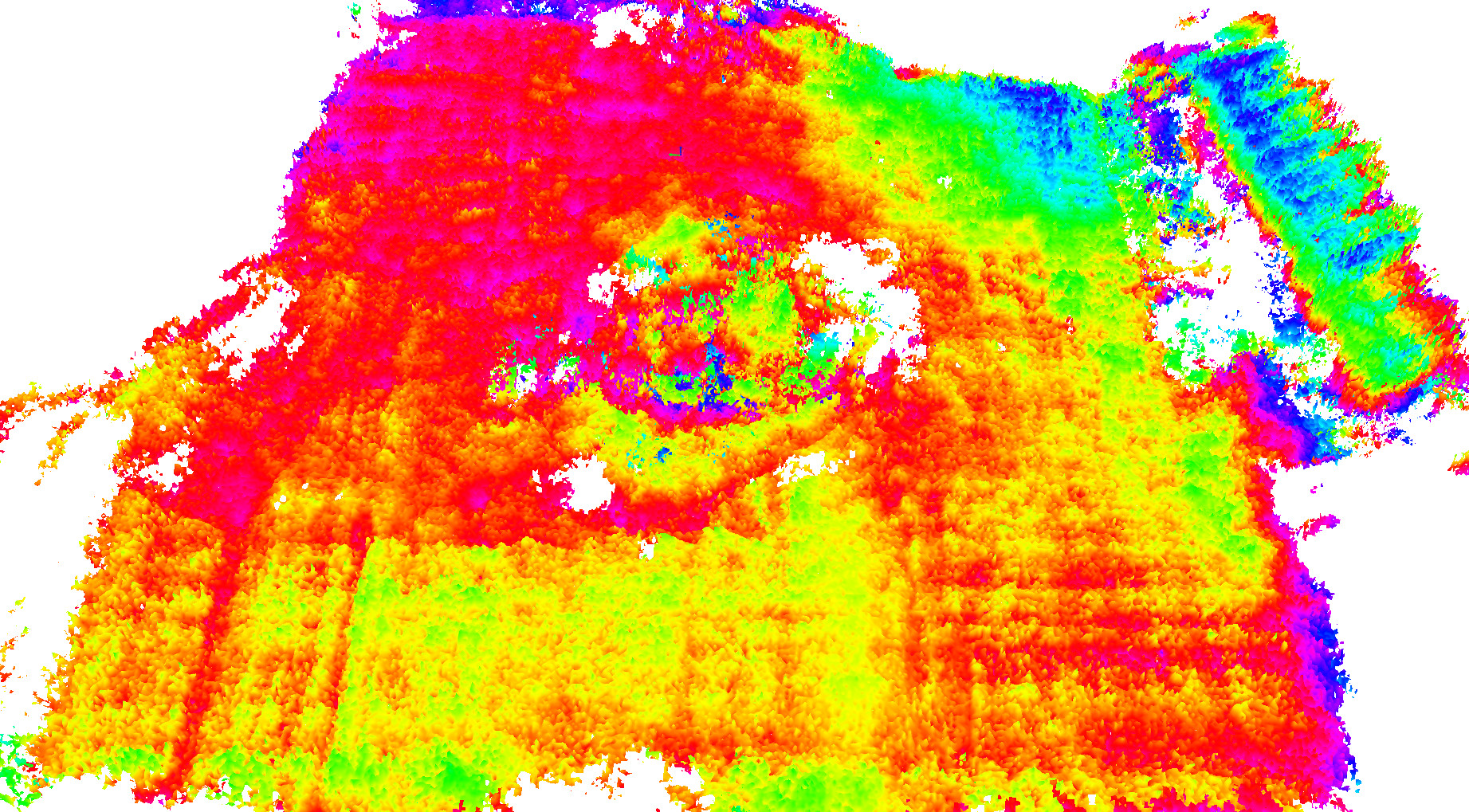}
  \caption{Second half of the square trajectory dataset: Polygon mesh reconstruction of the seafloor with a HSV color mapping repeating every 5 meters along the z axis.}
  \label{fig:square_4th_round_025}
\end{figure}

The reconstructed mesh of the second half of the square trajectory dataset is visualized in Fig. \ref{fig:square_4th_round_025}. It shows the 3D reconstruction of the seafloor with a HSV color mapping repeating every 5 meters along the z axis. The mesh was reconstructed with a grid cell resolution of $0.2 m$ and based only on the sonar measurements during the traverse of one square trajectory. In areas that have been covered from a variety of sonar viewpoints, the reconstruction shows to be consistent with the sparse DVL range measurements. For some cells, that have been covered by the sonar only close to a high azimuth angle and from very limited viewpoints, the models fail to classify the cells correctly, as can be seen in the center and the upper right part of the mesh.

\begin{figure}[t!]
  \centering
  \includegraphics [width=\columnwidth] {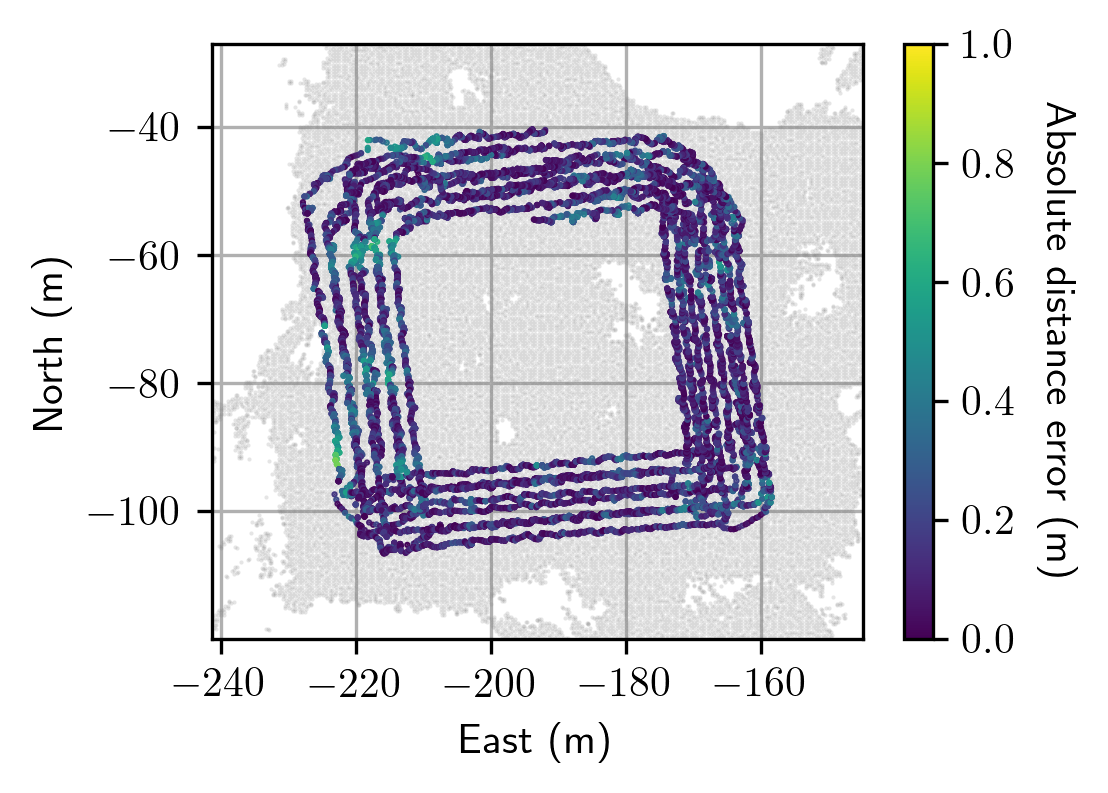}
  \caption{Square trajectory dataset: Absolute distance error in meter between the DVL range measurements and the reconstructed mesh. The outline of the mesh is illustrated in grey.}
  \label{fig:square_4th_round_025_eval}
\end{figure}

In order to evaluate the accuracy of the reconstructed mesh the closest distance between the sparse DVL range measurements and the mesh was computed. Fig. \ref{fig:square_4th_round_025_eval} shows the absolute distance error between the DVL range measurements and the reconstructed mesh with a MSE of $0.033 m^2$. The outline of the mesh is visible in the background.

\subsubsection{Low altitude dataset}
\begin{figure}[t!]
\vspace{5pt}
  \centering
  \includegraphics [width=\columnwidth] {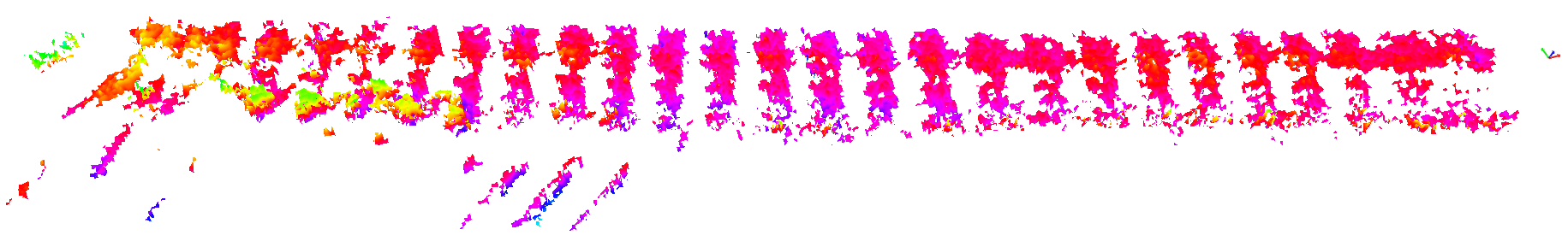}
  \caption{Low altitude dataset: Polygon mesh reconstruction of the seafloor with a HSV color mapping repeating every 5 meters along the z axis.}
  \label{fig:random_init_025}
\end{figure}

In a second dataset the vehicle followed a straight line for $100 m$ in a lower altitude ($2$  to $3 m$) to the seafloor.
The polygon mesh was reconstructed based on a cell resolution of $0.25 m$. Fig. \ref{fig:random_init_025} shows the mesh of the seafloor sensed by the imaging sonar with a HSV color mapping repeating every 5 meters along the z axis.

\begin{figure}[t!]
  \centering
  \includegraphics [width=\columnwidth] {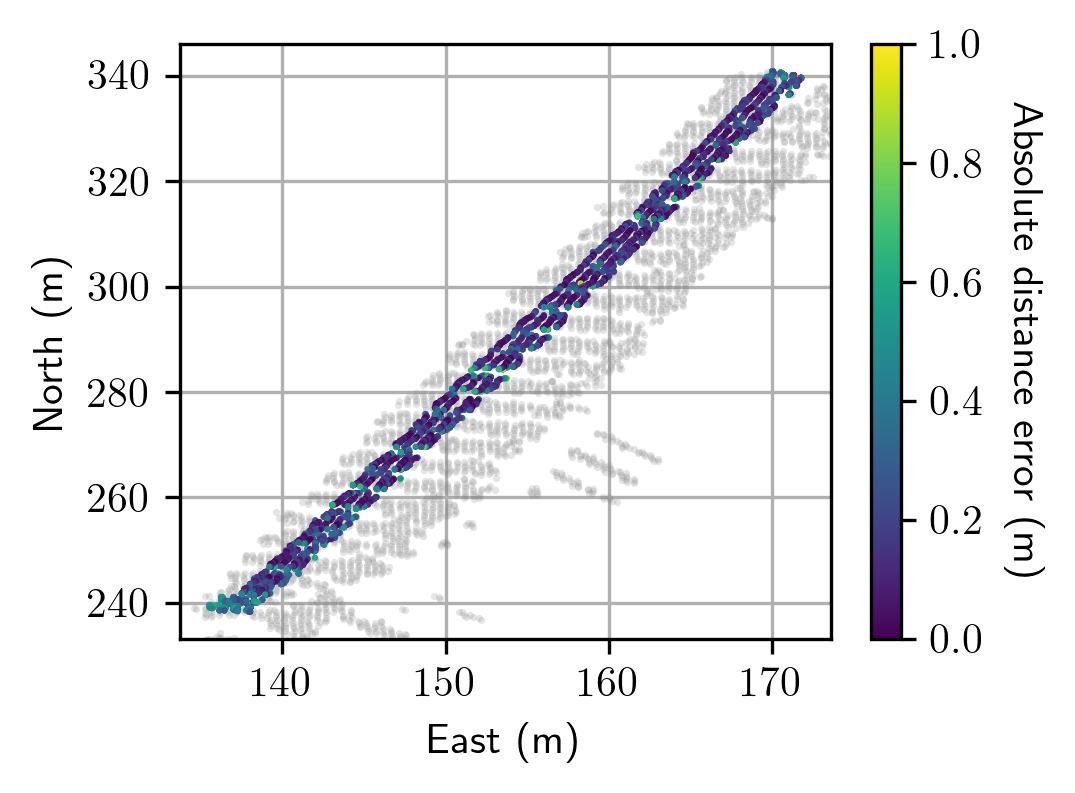}
  \caption{Low altitude dataset: Absolute distance error in meter between the DVL range measurements and the reconstructed mesh. The outline of the mesh is illustrated in grey.}
  \label{fig:random_init_025_eval}
\end{figure}

Fig. \ref{fig:random_init_025_eval} shows the absolute distance error between the DVL range measurements and the reconstructed mesh with a MSE of $0.067 m^2$. The mesh shows to be consistent with the ground truth, even though the models have been trained with higher altitude and on a different dataset. The mesh also shows regular gaps that are likely related to the pitch motion of the AUV.

\subsubsection{Man-made structures dataset}

\begin{figure}
  \vspace{5pt}
\centering
\begin{subfigure}{.5\columnwidth}
  \centering
  \includegraphics [width=\textwidth] {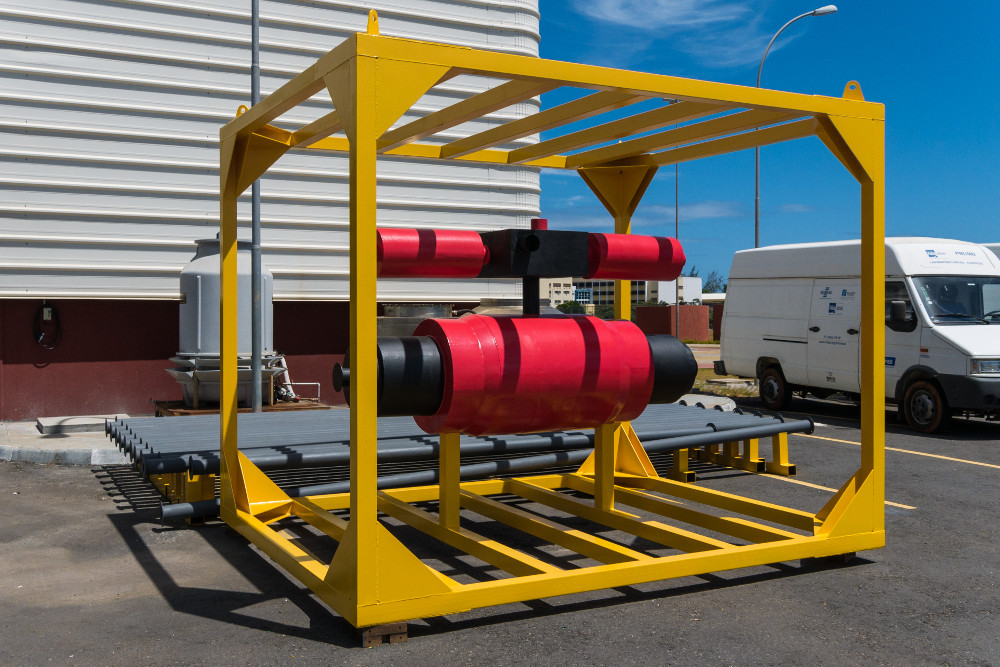}
  \caption{SSIV}
  \label{fig:man_made_structures_SSIV}
\end{subfigure}%
\begin{subfigure}{.5\columnwidth}
  \centering
  \includegraphics [width=\textwidth] {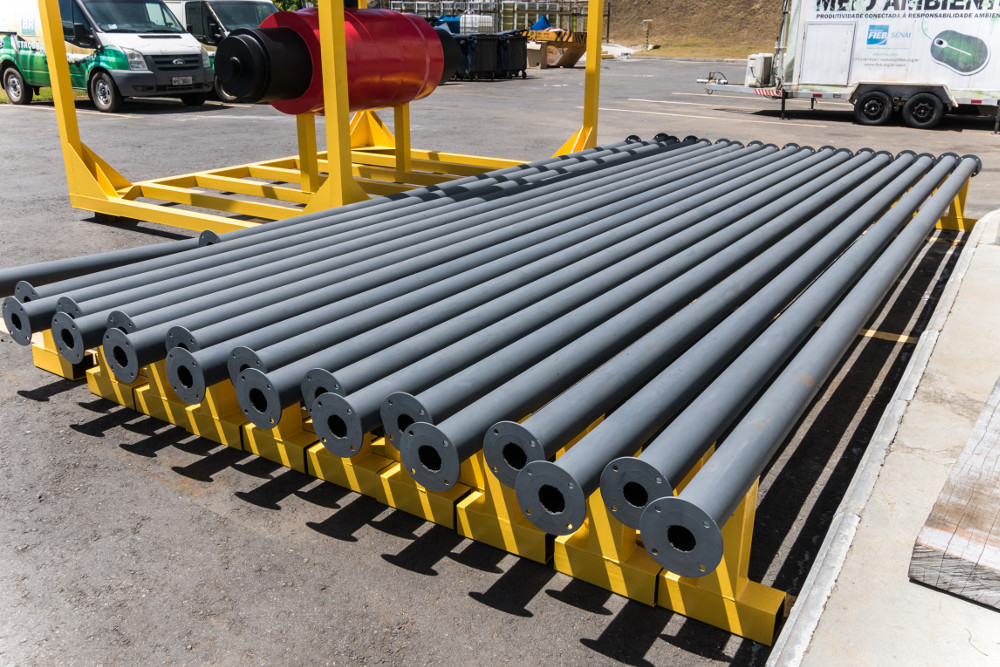}
  \caption{Pipeline elements}
  \label{fig:man_made_structures_pipeline}
\end{subfigure}
\caption{Subsea mockup structures. Images: Jan Albiez, SENAI CIMATEC}
\label{fig:man_made_structures}
\end{figure}

In order to evaluate if the model can also be applied to datasets, containing (vertical) structures that the model has not seen before, the model was applied to a third dataset containing mockups of a pipeline and a subsea isolation valve (SSIV). The mockups are shown in Fig. \ref{fig:man_made_structures}.

\begin{figure}[t!]
  \centering
  \def\svgwidth{1.\columnwidth}
  \small
  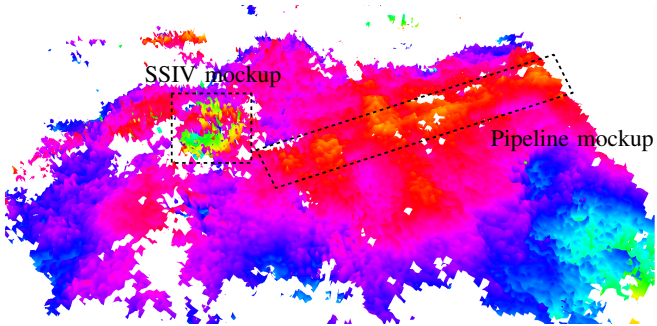
  \caption{Man-made structures dataset: Polygon mesh reconstruction of the seafloor sensed by the imaging sonar with a HSV color mapping repeating every 5 meters along the z axis.}
  \label{fig:ff_pipeline_02}
\end{figure}

In the reconstructed polygon mesh (Fig. \ref{fig:ff_pipeline_02}) the locations of the pipeline and the SSIV can be identified. The model is able to reconstruct some of the vertical surfaces of the $2.8 m$ high SSIV mockup.
However the shadows casted by the SSIV in the sonar image create gaps in the dense reconstruction close to the SSIV.

\begin{figure}[t!]
  \centering
  \includegraphics [width=\columnwidth] {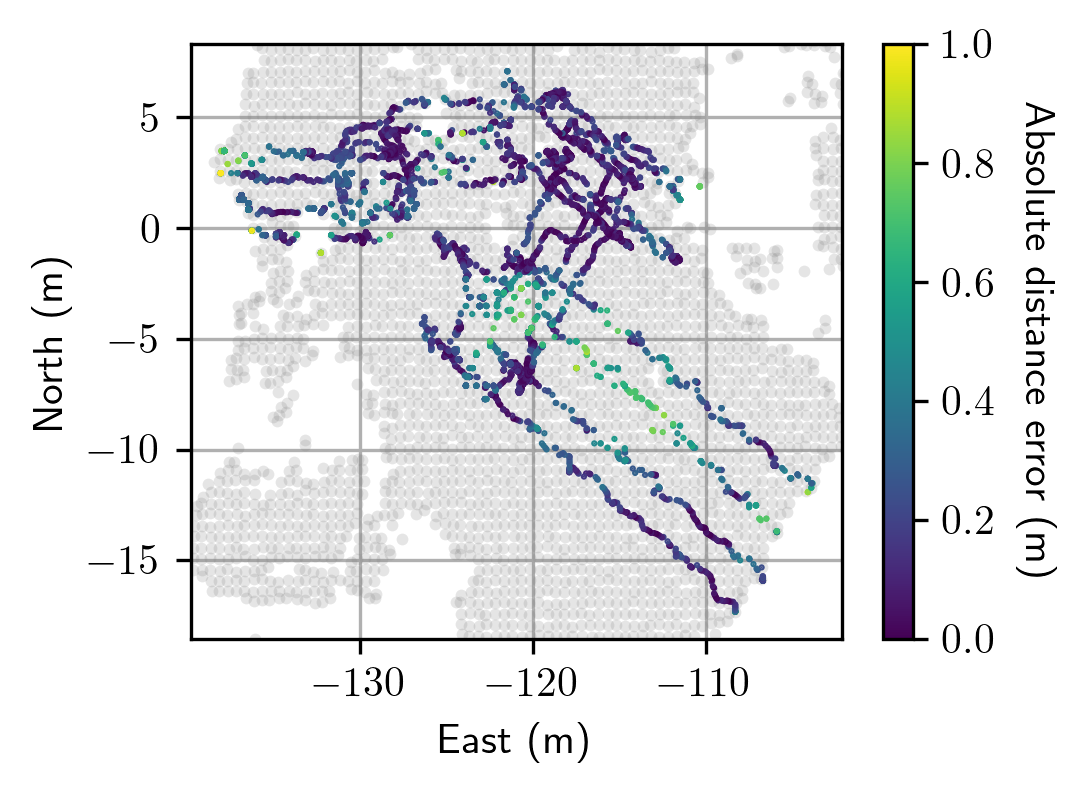}
  \caption{Man-made structures dataset: Absolute distance error in meter between the DVL range measurements and the reconstructed mesh. The outline of the mesh is illustrated in grey.}
  \label{fig:ff_pipeline_02_eval}
\end{figure}

Fig. \ref{fig:ff_pipeline_02_eval} shows the absolute distance error between the DVL range measurements and the reconstructed mesh with a MSE of $0.12 m^2$. In this dataset the mesh shows inconsistencies with the ground truth especially in the area of the pipeline.

\subsection{Different cell resolutions}
Since the models predict for each cell individually, they can be applied to different resolutions of the spatial acoustic feature grid while still being able to reconstruct the same consistent mesh. If the resolution is decreased of course the accuracy of the reconstruction suffers.
Table \ref{tab:cell_resolutions} gives an overview on different cell resolutions and the MSE of the reconstructed meshes for the square trajectory dataset. While the memory footprint grows cubically with the selected cell resolution, the MSE shows to be closer to linear improvement.

\begin{table}[]
\vspace{5pt}
\caption{Accuracy of the reconstructed meshes with respect to the cell resolution.}
\label{tab:cell_resolutions}
\centering
\begin{tabular}{|l|l|}
\hline
Cell resolution & \thead{MSE to ground truth} \\ \hline
$0.2m$  & $0.033 m^2$ \\ \hline
$0.25m$  & $0.038 m^2$ \\ \hline
$0.4m$  & $0.051 m^2$ \\ \hline
$0.6m$  & $0.064 m^2$ \\ \hline
\end{tabular}
\end{table}



\section{CONCLUSION}
\label{section:conclusion}

In this work we have presented a novel approach for the reconstruction of 3D surfaces using an imaging sonar sensor. We defined a feature vector which can be utilized to train models that can predict the signed distance and direction to the nearest surface in the scene.
The feature vector however could be replaced by another structure integrating the measured intensities from different viewpoints, for instance the discretized surface of a unit sphere. This would on the other hand also require a graph convolutional network architecture in order to model the neighborhood correctly.
One limitation of our current approach is that the predicted surface direction is represented as a delta on the unit sphere with respect to a fixed reference axis (unit z-axis). While for datasets that are focused on the bathymetry this seems sufficient, the reference could be selected uniquely for each cell using additional constraints.
Another limitation is that we disregard shadows and multi-path reflections present in the sonar image. These sections could be masked or modeled in order to improve the reconstruction result.
It is part of future work to compare the results of our work with existing solutions, like space carving.
We demonstrated the effectiveness of our approach on three real world datasets. While the sonar sensor in the datasets mainly imaged the seafloor, we could also show that man-made structures with vertical surfaces could be partially reconstructed by the models, even so they have not been present during the training step.

\addtolength{\textheight}{-8.5cm}   





\section*{ACKNOWLEDGMENT}

This work has been partially supported by the EurEx-LUNa project
(grant No. 50NA2002) funded by the German Federal Ministry of Economics and Technology (BMWi) and the H2020-ICT-2020-2 ICT-47-2020 project DeeperSense (Ref 101016958) funded by the European Union's Horizon 2020 research and innovation programme.


\bibliography{Bib}
\bibliographystyle{plain}

\end{document}

%% file: pics/sonar_image_side.pdf_tex
\begingroup%
  \makeatletter%
  \providecommand\color[2][]{%
    \errmessage{(Inkscape) Color is used for the text in Inkscape, but the package 'color.sty' is not loaded}%
    \renewcommand\color[2][]{}%
  }%
  \providecommand\transparent[1]{%
    \errmessage{(Inkscape) Transparency is used (non-zero) for the text in Inkscape, but the package 'transparent.sty' is not loaded}%
    \renewcommand\transparent[1]{}%
  }%
  \providecommand\rotatebox[2]{#2}%
  \newcommand*\fsize{\dimexpr\f@size pt\relax}%
  \newcommand*\lineheight[1]{\fontsize{\fsize}{#1\fsize}\selectfont}%
  \ifx\svgwidth\undefined%
    \setlength{\unitlength}{437.3495635bp}%
    \ifx\svgscale\undefined%
      \relax%
    \else%
      \setlength{\unitlength}{\unitlength * \real{\svgscale}}%
    \fi%
  \else%
    \setlength{\unitlength}{\svgwidth}%
  \fi%
  \global\let\svgwidth\undefined%
  \global\let\svgscale\undefined%
  \makeatother%
  \begin{picture}(1,0.51862506)%
    \lineheight{1}%
    \setlength\tabcolsep{0pt}%
    \put(2.15759488,-0.4034845){\color[rgb]{0,0,0}\makebox(0,0)[lt]{\begin{minipage}{2.09257237\unitlength}\raggedright \end{minipage}}}%
    \put(0,0){\includegraphics[width=\unitlength,page=1]{sonar_image_side.pdf}}%
    \put(0.4858863,0.34728075){\color[rgb]{0,0,0}\makebox(0,0)[lt]{\lineheight{1.25}\smash{\begin{tabular}[t]{l}$r$\end{tabular}}}}%
    \put(0.49365508,0.49777861){\color[rgb]{0,0,0}\makebox(0,0)[lt]{\lineheight{1.25}\smash{\begin{tabular}[t]{l}$\phi_{\textrm{min}}$\end{tabular}}}}%
    \put(0.4992569,0.00647097){\color[rgb]{0,0,0}\makebox(0,0)[lt]{\lineheight{1.25}\smash{\begin{tabular}[t]{l}$\phi_{\textrm{max}}$\end{tabular}}}}%
    \put(0.11194243,0.35632784){\color[rgb]{0,0,0}\makebox(0,0)[lt]{\lineheight{1.25}\smash{\begin{tabular}[t]{l}$\phi$\end{tabular}}}}%
    \put(0.71161959,0.24769989){\color[rgb]{0,0,0}\makebox(0,0)[lt]{\lineheight{1.25}\smash{\begin{tabular}[t]{l}$r_{\textrm{max}}$\end{tabular}}}}%
    \put(0,0){\includegraphics[width=\unitlength,page=2]{sonar_image_side.pdf}}%
    \put(0.02430306,0.4081627){\color[rgb]{0,0,0}\makebox(0,0)[lt]{\begin{minipage}{0.03268453\unitlength}\raggedright \end{minipage}}}%
    \put(0.02240137,0.39673584){\color[rgb]{0,0,0}\makebox(0,0)[lt]{\lineheight{1.25}\smash{\begin{tabular}[t]{l}$z$\end{tabular}}}}%
    \put(0.12885948,0.21776091){\color[rgb]{0,0,0}\makebox(0,0)[lt]{\lineheight{1.25}\smash{\begin{tabular}[t]{l}$x$\end{tabular}}}}%
    \put(0,0){\includegraphics[width=\unitlength,page=3]{sonar_image_side.pdf}}%
  \end{picture}%
\endgroup%

%% file: pics/sonar_image_above.pdf_tex
\begingroup%
  \makeatletter%
  \providecommand\color[2][]{%
    \errmessage{(Inkscape) Color is used for the text in Inkscape, but the package 'color.sty' is not loaded}%
    \renewcommand\color[2][]{}%
  }%
  \providecommand\transparent[1]{%
    \errmessage{(Inkscape) Transparency is used (non-zero) for the text in Inkscape, but the package 'transparent.sty' is not loaded}%
    \renewcommand\transparent[1]{}%
  }%
  \providecommand\rotatebox[2]{#2}%
  \newcommand*\fsize{\dimexpr\f@size pt\relax}%
  \newcommand*\lineheight[1]{\fontsize{\fsize}{#1\fsize}\selectfont}%
  \ifx\svgwidth\undefined%
    \setlength{\unitlength}{597.77887819bp}%
    \ifx\svgscale\undefined%
      \relax%
    \else%
      \setlength{\unitlength}{\unitlength * \real{\svgscale}}%
    \fi%
  \else%
    \setlength{\unitlength}{\svgwidth}%
  \fi%
  \global\let\svgwidth\undefined%
  \global\let\svgscale\undefined%
  \makeatother%
  \begin{picture}(1,0.54441311)%
    \lineheight{1}%
    \setlength\tabcolsep{0pt}%
    \put(1.75576166,-0.22457831){\color[rgb]{0,0,0}\makebox(0,0)[lt]{\begin{minipage}{1.53097683\unitlength}\raggedright \end{minipage}}}%
    \put(0,0){\includegraphics[width=\unitlength,page=1]{sonar_image_above.pdf}}%
    \put(0.26749166,0.35766707){\color[rgb]{0,0,0}\makebox(0,0)[lt]{\lineheight{1.25}\smash{\begin{tabular}[t]{l}$r$\end{tabular}}}}%
    \put(0.74321937,0.1321818){\color[rgb]{0,0,0}\makebox(0,0)[lt]{\lineheight{1.25}\smash{\begin{tabular}[t]{l}$\theta_{\textrm{min}}$\end{tabular}}}}%
    \put(0.07820146,0.13158269){\color[rgb]{0,0,0}\makebox(0,0)[lt]{\lineheight{1.25}\smash{\begin{tabular}[t]{l}$\theta_{\textrm{max}}$\end{tabular}}}}%
    \put(0.43736953,0.52916134){\color[rgb]{0,0,0}\makebox(0,0)[lt]{\lineheight{1.25}\smash{\begin{tabular}[t]{l}$r_{\textrm{max}}$\end{tabular}}}}%
    \put(0,0){\includegraphics[width=\unitlength,page=2]{sonar_image_above.pdf}}%
    \put(0.19499349,0.36924252){\color[rgb]{0,0,0}\makebox(0,0)[lt]{\begin{minipage}{0.02391279\unitlength}\raggedright \end{minipage}}}%
    \put(0.31897285,0.02374947){\color[rgb]{0,0,0}\makebox(0,0)[lt]{\lineheight{1.25}\smash{\begin{tabular}[t]{l}$y$\end{tabular}}}}%
    \put(0.45592704,0.07856634){\color[rgb]{0,0,0}\makebox(0,0)[lt]{\lineheight{1.25}\smash{\begin{tabular}[t]{l}$x$\end{tabular}}}}%
    \put(0,0){\includegraphics[width=\unitlength,page=3]{sonar_image_above.pdf}}%
    \put(0.36952137,0.3501385){\color[rgb]{0,0,0}\makebox(0,0)[lt]{\lineheight{1.25}\smash{\begin{tabular}[t]{l}$\theta$\end{tabular}}}}%
  \end{picture}%
\endgroup%

%% file: pics/spatial_acoustic_reconstruction_annotated.pdf_tex
\begingroup%
  \makeatletter%
  \providecommand\color[2][]{%
    \errmessage{(Inkscape) Color is used for the text in Inkscape, but the package 'color.sty' is not loaded}%
    \renewcommand\color[2][]{}%
  }%
  \providecommand\transparent[1]{%
    \errmessage{(Inkscape) Transparency is used (non-zero) for the text in Inkscape, but the package 'transparent.sty' is not loaded}%
    \renewcommand\transparent[1]{}%
  }%
  \providecommand\rotatebox[2]{#2}%
  \newcommand*\fsize{\dimexpr\f@size pt\relax}%
  \newcommand*\lineheight[1]{\fontsize{\fsize}{#1\fsize}\selectfont}%
  \ifx\svgwidth\undefined%
    \setlength{\unitlength}{440.5220274bp}%
    \ifx\svgscale\undefined%
      \relax%
    \else%
      \setlength{\unitlength}{\unitlength * \real{\svgscale}}%
    \fi%
  \else%
    \setlength{\unitlength}{\svgwidth}%
  \fi%
  \global\let\svgwidth\undefined%
  \global\let\svgscale\undefined%
  \makeatother%
  \begin{picture}(1,0.59434294)%
    \lineheight{1}%
    \setlength\tabcolsep{0pt}%
    \put(0,0){\includegraphics[width=\unitlength,page=1]{spatial_acoustic_reconstruction_annotated.pdf}}%
    \put(0.30328807,0.59672764){\color[rgb]{0,0,0}\makebox(0,0)[lt]{\begin{minipage}{0.16830681\unitlength}\raggedright training\end{minipage}}}%
    \put(0.68379464,0.59774726){\color[rgb]{0,0,0}\makebox(0,0)[lt]{\begin{minipage}{0.16830682\unitlength}\raggedright prediction\end{minipage}}}%
  \end{picture}%
\endgroup%

%% file: pics/inverse_projection.pdf_tex
\begingroup%
  \makeatletter%
  \providecommand\color[2][]{%
    \errmessage{(Inkscape) Color is used for the text in Inkscape, but the package 'color.sty' is not loaded}%
    \renewcommand\color[2][]{}%
  }%
  \providecommand\transparent[1]{%
    \errmessage{(Inkscape) Transparency is used (non-zero) for the text in Inkscape, but the package 'transparent.sty' is not loaded}%
    \renewcommand\transparent[1]{}%
  }%
  \providecommand\rotatebox[2]{#2}%
  \newcommand*\fsize{\dimexpr\f@size pt\relax}%
  \newcommand*\lineheight[1]{\fontsize{\fsize}{#1\fsize}\selectfont}%
  \ifx\svgwidth\undefined%
    \setlength{\unitlength}{522.34849116bp}%
    \ifx\svgscale\undefined%
      \relax%
    \else%
      \setlength{\unitlength}{\unitlength * \real{\svgscale}}%
    \fi%
  \else%
    \setlength{\unitlength}{\svgwidth}%
  \fi%
  \global\let\svgwidth\undefined%
  \global\let\svgscale\undefined%
  \makeatother%
  \begin{picture}(1,0.57793964)%
    \lineheight{1}%
    \setlength\tabcolsep{0pt}%
    \put(0,0){\includegraphics[width=\unitlength,page=1]{inverse_projection.pdf}}%
    \put(0.26461551,0.21495837){\color[rgb]{0,0,0}\makebox(0,0)[lt]{\lineheight{1.25}\smash{\begin{tabular}[t]{l}$\mathbf{p}_c$\end{tabular}}}}%
    \put(0.53398912,0.55977836){\color[rgb]{0,0,0}\makebox(0,0)[lt]{\lineheight{1.25}\smash{\begin{tabular}[t]{l}$\mathbf{p}_{s,t+1}$\end{tabular}}}}%
    \put(0.65469285,0.56048542){\color[rgb]{0,0,0}\makebox(0,0)[lt]{\lineheight{1.25}\smash{\begin{tabular}[t]{l}$\mathbf{p}_{s,t}$\end{tabular}}}}%
    \put(0.74634167,0.5601172){\color[rgb]{0,0,0}\makebox(0,0)[lt]{\lineheight{1.25}\smash{\begin{tabular}[t]{l}$\mathbf{p}_{s,t-1}$\end{tabular}}}}%
    \put(0.47910584,0.37248719){\color[rgb]{0,0,0}\makebox(0,0)[lt]{\lineheight{1.25}\smash{\begin{tabular}[t]{l}$\mathbf{\hat{r}}_t$\end{tabular}}}}%
    \put(0.47230922,0.31293696){\color[rgb]{0,0,0}\makebox(0,0)[lt]{\lineheight{1.25}\smash{\begin{tabular}[t]{l}$\mathbf{\hat{r}}_{t-1}$\end{tabular}}}}%
    \put(0.40735458,0.41461446){\color[rgb]{0,0,0}\makebox(0,0)[lt]{\lineheight{1.25}\smash{\begin{tabular}[t]{l}$\mathbf{\hat{r}}_{t+1}$\end{tabular}}}}%
    \put(0.14665085,0.3876982){\color[rgb]{0,0,0}\makebox(0,0)[lt]{\lineheight{1.25}\smash{\begin{tabular}[t]{l}$I_{c,t+1}$\end{tabular}}}}%
    \put(0.25402847,0.31540779){\color[rgb]{0,0,0}\makebox(0,0)[lt]{\lineheight{1.25}\smash{\begin{tabular}[t]{l}$I_{c,t}$\end{tabular}}}}%
    \put(0.47244456,0.05048414){\color[rgb]{0,0,0}\makebox(0,0)[lt]{\lineheight{1.25}\smash{\begin{tabular}[t]{l}$I_{c,t-1}$\end{tabular}}}}%
  \end{picture}%
\endgroup%

%% file: pics/classifier_network.pdf_tex
\begingroup%
  \makeatletter%
  \providecommand\color[2][]{%
    \errmessage{(Inkscape) Color is used for the text in Inkscape, but the package 'color.sty' is not loaded}%
    \renewcommand\color[2][]{}%
  }%
  \providecommand\transparent[1]{%
    \errmessage{(Inkscape) Transparency is used (non-zero) for the text in Inkscape, but the package 'transparent.sty' is not loaded}%
    \renewcommand\transparent[1]{}%
  }%
  \providecommand\rotatebox[2]{#2}%
  \newcommand*\fsize{\dimexpr\f@size pt\relax}%
  \newcommand*\lineheight[1]{\fontsize{\fsize}{#1\fsize}\selectfont}%
  \ifx\svgwidth\undefined%
    \setlength{\unitlength}{922.86277651bp}%
    \ifx\svgscale\undefined%
      \relax%
    \else%
      \setlength{\unitlength}{\unitlength * \real{\svgscale}}%
    \fi%
  \else%
    \setlength{\unitlength}{\svgwidth}%
  \fi%
  \global\let\svgwidth\undefined%
  \global\let\svgscale\undefined%
  \makeatother%
  \begin{picture}(1,0.62782195)%
    \lineheight{1}%
    \setlength\tabcolsep{0pt}%
    \put(0,0){\includegraphics[width=\unitlength,page=1]{classifier_network.pdf}}%
    \put(0.00517905,0.33103731){\color[rgb]{0,0,0}\rotatebox{-90}{\makebox(0,0)[lt]{\lineheight{1.25}\smash{\begin{tabular}[t]{l}Input($3n$,1)\end{tabular}}}}}%
    \put(0.06170006,0.33069327){\color[rgb]{0,0,0}\rotatebox{-90}{\makebox(0,0)[lt]{\lineheight{1.25}\smash{\begin{tabular}[t]{l}Normalization\end{tabular}}}}}%
    \put(0,0){\includegraphics[width=\unitlength,page=2]{classifier_network.pdf}}%
    \put(0.8346007,0.3307164){\color[rgb]{0,0,0}\rotatebox{-90}{\makebox(0,0)[lt]{\lineheight{1.25}\smash{\begin{tabular}[t]{l}GlobalAveragePooling1D\end{tabular}}}}}%
    \put(0.89148865,0.33071645){\color[rgb]{0,0,0}\rotatebox{-90}{\makebox(0,0)[lt]{\lineheight{1.25}\smash{\begin{tabular}[t]{l}Dense(96)\end{tabular}}}}}%
    \put(0.94024996,0.33071645){\color[rgb]{0,0,0}\rotatebox{-90}{\makebox(0,0)[lt]{\lineheight{1.25}\smash{\begin{tabular}[t]{l}Dropout(0.1)\end{tabular}}}}}%
    \put(0.98901127,0.3307164){\color[rgb]{0,0,0}\rotatebox{-90}{\makebox(0,0)[lt]{\lineheight{1.25}\smash{\begin{tabular}[t]{l}Dense(1)\end{tabular}}}}}%
    \put(0.13568863,0.33071518){\color[rgb]{0,0,0}\rotatebox{-90}{\makebox(0,0)[lt]{\lineheight{1.25}\smash{\begin{tabular}[t]{l}Conv1D(16,5)\end{tabular}}}}}%
    \put(0.2169575,0.33071518){\color[rgb]{0,0,0}\rotatebox{-90}{\makebox(0,0)[lt]{\lineheight{1.25}\smash{\begin{tabular}[t]{l}AveragePooling1D(2)\end{tabular}}}}}%
    \put(0.28847402,0.33071525){\color[rgb]{0,0,0}\rotatebox{-90}{\makebox(0,0)[lt]{\lineheight{1.25}\smash{\begin{tabular}[t]{l}Conv1D(32,5)\end{tabular}}}}}%
    \put(0.37136821,0.32908973){\color[rgb]{0,0,0}\rotatebox{-90}{\makebox(0,0)[lt]{\lineheight{1.25}\smash{\begin{tabular}[t]{l}AveragePooling1D(2)\end{tabular}}}}}%
    \put(0.44450927,0.33071511){\color[rgb]{0,0,0}\rotatebox{-90}{\makebox(0,0)[lt]{\lineheight{1.25}\smash{\begin{tabular}[t]{l}Conv1D(48,5)\end{tabular}}}}}%
    \put(0.52577801,0.33071511){\color[rgb]{0,0,0}\rotatebox{-90}{\makebox(0,0)[lt]{\lineheight{1.25}\smash{\begin{tabular}[t]{l}AveragePooling1D(2)\end{tabular}}}}}%
    \put(0.59891993,0.33071511){\color[rgb]{0,0,0}\rotatebox{-90}{\makebox(0,0)[lt]{\lineheight{1.25}\smash{\begin{tabular}[t]{l}Conv1D(64,5)\end{tabular}}}}}%
    \put(0.68018872,0.33071511){\color[rgb]{0,0,0}\rotatebox{-90}{\makebox(0,0)[lt]{\lineheight{1.25}\smash{\begin{tabular}[t]{l}AveragePooling1D(2)\end{tabular}}}}}%
    \put(0.75333064,0.33071511){\color[rgb]{0,0,0}\rotatebox{-90}{\makebox(0,0)[lt]{\lineheight{1.25}\smash{\begin{tabular}[t]{l}Conv1D(80,5)\end{tabular}}}}}%
    \put(0,0){\includegraphics[width=\unitlength,page=3]{classifier_network.pdf}}%
  \end{picture}%
\endgroup%

%% file: pics/regressor_network.pdf_tex
\begingroup%
  \makeatletter%
  \providecommand\color[2][]{%
    \errmessage{(Inkscape) Color is used for the text in Inkscape, but the package 'color.sty' is not loaded}%
    \renewcommand\color[2][]{}%
  }%
  \providecommand\transparent[1]{%
    \errmessage{(Inkscape) Transparency is used (non-zero) for the text in Inkscape, but the package 'transparent.sty' is not loaded}%
    \renewcommand\transparent[1]{}%
  }%
  \providecommand\rotatebox[2]{#2}%
  \newcommand*\fsize{\dimexpr\f@size pt\relax}%
  \newcommand*\lineheight[1]{\fontsize{\fsize}{#1\fsize}\selectfont}%
  \ifx\svgwidth\undefined%
    \setlength{\unitlength}{922.89534152bp}%
    \ifx\svgscale\undefined%
      \relax%
    \else%
      \setlength{\unitlength}{\unitlength * \real{\svgscale}}%
    \fi%
  \else%
    \setlength{\unitlength}{\svgwidth}%
  \fi%
  \global\let\svgwidth\undefined%
  \global\let\svgscale\undefined%
  \makeatother%
  \begin{picture}(1,0.60350297)%
    \lineheight{1}%
    \setlength\tabcolsep{0pt}%
    \put(0,0){\includegraphics[width=\unitlength,page=1]{regressor_network.pdf}}%
    \put(0.00517887,0.30177003){\color[rgb]{0,0,0}\rotatebox{-90}{\makebox(0,0)[lt]{\lineheight{1.25}\smash{\begin{tabular}[t]{l}Input($3n$,1)\end{tabular}}}}}%
    \put(0.06169789,0.301426){\color[rgb]{0,0,0}\rotatebox{-90}{\makebox(0,0)[lt]{\lineheight{1.25}\smash{\begin{tabular}[t]{l}Normalization\end{tabular}}}}}%
    \put(0.13568377,0.30144923){\color[rgb]{0,0,0}\rotatebox{-90}{\makebox(0,0)[lt]{\lineheight{1.25}\smash{\begin{tabular}[t]{l}Conv1D(16,5)\end{tabular}}}}}%
    \put(0,0){\includegraphics[width=\unitlength,page=2]{regressor_network.pdf}}%
    \put(0.21694977,0.30144923){\color[rgb]{0,0,0}\rotatebox{-90}{\makebox(0,0)[lt]{\lineheight{1.25}\smash{\begin{tabular}[t]{l}AveragePooling1D(2)\end{tabular}}}}}%
    \put(0.28846377,0.3014493){\color[rgb]{0,0,0}\rotatebox{-90}{\makebox(0,0)[lt]{\lineheight{1.25}\smash{\begin{tabular}[t]{l}Conv1D(32,5)\end{tabular}}}}}%
    \put(0.37135518,0.29982384){\color[rgb]{0,0,0}\rotatebox{-90}{\makebox(0,0)[lt]{\lineheight{1.25}\smash{\begin{tabular}[t]{l}AveragePooling1D(2)\end{tabular}}}}}%
    \put(0.44449442,0.30144913){\color[rgb]{0,0,0}\rotatebox{-90}{\makebox(0,0)[lt]{\lineheight{1.25}\smash{\begin{tabular}[t]{l}Conv1D(48,5)\end{tabular}}}}}%
    \put(0.52576044,0.30144913){\color[rgb]{0,0,0}\rotatebox{-90}{\makebox(0,0)[lt]{\lineheight{1.25}\smash{\begin{tabular}[t]{l}AveragePooling1D(2)\end{tabular}}}}}%
    \put(0.59889978,0.30144913){\color[rgb]{0,0,0}\rotatebox{-90}{\makebox(0,0)[lt]{\lineheight{1.25}\smash{\begin{tabular}[t]{l}Conv1D(64,5)\end{tabular}}}}}%
    \put(0.68016571,0.30144913){\color[rgb]{0,0,0}\rotatebox{-90}{\makebox(0,0)[lt]{\lineheight{1.25}\smash{\begin{tabular}[t]{l}AveragePooling1D(2)\end{tabular}}}}}%
    \put(0.79393651,0.30144918){\color[rgb]{0,0,0}\rotatebox{-90}{\makebox(0,0)[lt]{\lineheight{1.25}\smash{\begin{tabular}[t]{l}Dense(80)\end{tabular}}}}}%
    \put(0.94021678,0.30144918){\color[rgb]{0,0,0}\rotatebox{-90}{\makebox(0,0)[lt]{\lineheight{1.25}\smash{\begin{tabular}[t]{l}Dropout(0.1)\end{tabular}}}}}%
    \put(0.98897637,0.30144913){\color[rgb]{0,0,0}\rotatebox{-90}{\makebox(0,0)[lt]{\lineheight{1.25}\smash{\begin{tabular}[t]{l}Dense(3)\end{tabular}}}}}%
    \put(0,0){\includegraphics[width=\unitlength,page=3]{regressor_network.pdf}}%
    \put(0.73705062,0.30144923){\color[rgb]{0,0,0}\rotatebox{-90}{\makebox(0,0)[lt]{\lineheight{1.25}\smash{\begin{tabular}[t]{l}Flatten\end{tabular}}}}}%
    \put(0,0){\includegraphics[width=\unitlength,page=4]{regressor_network.pdf}}%
    \put(0.84269694,0.30144913){\color[rgb]{0,0,0}\rotatebox{-90}{\makebox(0,0)[lt]{\lineheight{1.25}\smash{\begin{tabular}[t]{l}Dropout(0.1)\end{tabular}}}}}%
    \put(0,0){\includegraphics[width=\unitlength,page=5]{regressor_network.pdf}}%
    \put(0.8914556,0.30144918){\color[rgb]{0,0,0}\rotatebox{-90}{\makebox(0,0)[lt]{\lineheight{1.25}\smash{\begin{tabular}[t]{l}Dense(32)\end{tabular}}}}}%
    \put(0,0){\includegraphics[width=\unitlength,page=6]{regressor_network.pdf}}%
  \end{picture}%
\endgroup%

%% file: pics/ff_pipeline_02.pdf_tex
\begingroup%
  \makeatletter%
  \providecommand\color[2][]{%
    \errmessage{(Inkscape) Color is used for the text in Inkscape, but the package 'color.sty' is not loaded}%
    \renewcommand\color[2][]{}%
  }%
  \providecommand\transparent[1]{%
    \errmessage{(Inkscape) Transparency is used (non-zero) for the text in Inkscape, but the package 'transparent.sty' is not loaded}%
    \renewcommand\transparent[1]{}%
  }%
  \providecommand\rotatebox[2]{#2}%
  \newcommand*\fsize{\dimexpr\f@size pt\relax}%
  \newcommand*\lineheight[1]{\fontsize{\fsize}{#1\fsize}\selectfont}%
  \ifx\svgwidth\undefined%
    \setlength{\unitlength}{1843.99998462bp}%
    \ifx\svgscale\undefined%
      \relax%
    \else%
      \setlength{\unitlength}{\unitlength * \real{\svgscale}}%
    \fi%
  \else%
    \setlength{\unitlength}{\svgwidth}%
  \fi%
  \global\let\svgwidth\undefined%
  \global\let\svgscale\undefined%
  \makeatother%
  \begin{picture}(1,0.49078096)%
    \lineheight{1}%
    \setlength\tabcolsep{0pt}%
    \put(0,0){\includegraphics[width=\unitlength,page=1]{ff_pipeline_02.pdf}}%
    \put(0.74755219,0.2983439){\color[rgb]{0,0,0}\makebox(0,0)[lt]{\begin{minipage}{0.2593792\unitlength}\raggedright Pipeline mockup\end{minipage}}}%
    \put(0.21487352,0.39755023){\color[rgb]{0,0,0}\makebox(0,0)[lt]{\begin{minipage}{0.28877445\unitlength}\raggedright SSIV mockup\end{minipage}}}%
  \end{picture}%
\endgroup%